\pdfoutput=1

\documentclass[11pt]{article}
\usepackage[]{acl}
\usepackage{graphicx}
\usepackage{times}
\usepackage{latexsym}
\usepackage{amsmath}

\usepackage[T1]{fontenc}

\usepackage[utf8]{inputenc}

\usepackage{microtype}
\makeatletter

\renewcommand{\maketag@@@}[1]{\hbox{\m@th\normalsize\normalfont#1}}%

\title{OVEL: Large Language Model as Memory Manager for Online Video Entity Linking}

\author{
    Haiquan Zhao\textsuperscript{1}, 
    Xuwu Wang\textsuperscript{1}, 
    Shisong Chen\textsuperscript{2}, \\
    \bf {Zhixu Li\textsuperscript{1}\thanks{~~Corresponding author},
        Xin Zheng\textsuperscript{3}, 
        Yanghua Xiao\textsuperscript{1}\footnotemark[1]} \\
    \textsuperscript{1}Shanghai Key Laboratory of Data Science, School of Computer Science, Fudan University \\
    \textsuperscript{2} Shanghai Institute of AI for Education and School of Computer Science and Technology,\\ East China Normal University
    \textsuperscript{3}iFLYTEK CO., LTD, Suzhou, China \\
    \texttt{zhaohq22@m.fudan.edu.cn},~\texttt{\{xuwang18,zhixuli,shawyh\}@fudan.edu.cn} ,\\~ \texttt{sschen@stu.ecnu.edu.cn}, 
    ~\texttt{xinzheng3@iflytek.com}
}
\begin{document}
\maketitle
\begin{abstract}
In recent years, multi-modal entity linking (MEL) has garnered increasing attention in the research community due to its significance in numerous multi-modal applications. Video, as a popular means of information transmission, has become prevalent in people's daily lives. However, most existing MEL methods primarily focus on linking textual and visual mentions or offline videos's mentions to entities in multi-modal knowledge bases, with limited efforts devoted to linking mentions within online video content. In this paper, we propose a task called Online Video Entity Linking (\textit{OVEL}), aiming to establish connections between mentions in online videos and a knowledge base with high accuracy and timeliness. To facilitate the research works of \textit{OVEL}, we specifically concentrate on live delivery scenarios
and construct a live delivery entity linking dataset called \textit{LIVE}. Besides, we propose an evaluation metric 
that considers timelessness, robustness, and accuracy. Furthermore, to effectively handle \textit{OVEL} task, we leverage a memory block managed by a Large Language Model
and retrieve entity candidates from the knowledge base to augment LLM performance on memory management. The experimental results prove the effectiveness and efficiency of our method. 
\end{abstract}

\section{Introduction}
\begin{figure}[ht]
\centering
\includegraphics[width=0.5\textwidth]{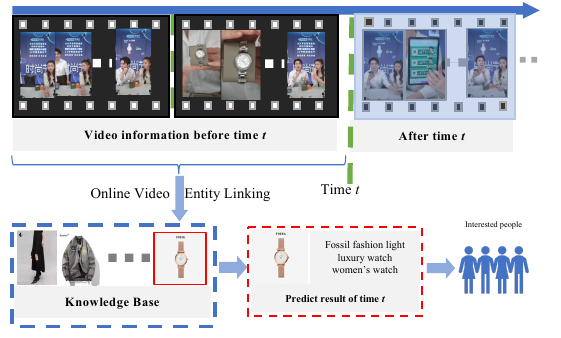}
\caption{The task of \textit{OVEL} in the live delivery scene. The upper represents an online delivery video. At time \textit{t}, 
it takes information before time \textit{t} as input and identifies salient entities from the video. Relevant entities are pushed to specific persons for recommendation.}
  \label{FIntro}
\end{figure}
Videos, showcased by platforms like TikTok and YouTube, have become a dominant medium for communication. As their significance grows, so does the breadth of academic research into understanding them. Beyond the well-studied areas of video retrieval and captioning, scholars\cite{xu2016msr,miech2019howto100m,gabeur2020multi,gan2021multimodal} are exploring aspects like pre-training, cross-modal fusion, and more, striving for a comprehensive grasp of video content.
However, these existing studies mainly concentrate on understanding the holistic content of videos and often overlook the significance of specific entities within them. Consider a live streaming example: where a video captioning model might merely state ``a host explaining a product'', however, for viewers, specific details like ``Nike Air Jordan 37th Generation Mid-Top Basketball Shoes'' might be the critical information they seek. Therefore, in such scenarios, discerning specific entities can be more vital than a broad overview of the video content. 

Video entity linking refers to linking mentions that appear in a video to their corresponding entities in a knowledge base. 
Related research on this task is still relatively limited. 
There have been studies\cite{adjali2020building,adjali2020multimodal,zhou2021weibo,wang2022wikidiverse,wang2022multimodal,gan2021multimodal,sun2022visual,luo2023multi,xing2023drin} focused on the research of Multimodal Entity Linking (MEL), which aims to link mentions of multiple modalities (primarily text and images) to a knowledge base. These works primarily focus on static visual-textual pairs, with limited consideration for mentions in video data. 
Besides, some studies \cite{li2015semantic,venkitasubramanian2017entity} have conducted video entity linking, but with certain limitations. On the one hand, they link to coarse-grained entities like ``bird'' or ``human'', which becomes overly simplistic due to the broad granularity. On the other hand, they don't demand real-time processing. With the rise of network terminals, there is an increasing demand for improved online performance in certain scenarios. For instance, in online sports live broadcasts, if specific athletes can be identified, comments and even real-time explanations can be generated based on the career of the athletes. These scenarios put forward higher requirements for online video entity linking. 

In this paper, we propose the task of Online Video Entity Linking (\textit{OVEL}) on dynamic video streams. The objective of this task is to link important entities appearing in online videos to a corresponding knowledge base. Furthermore, to advance the research on \textit{OVEL}, we construct a dataset for LIVE stream product recognition based on live streaming scenarios called \textit{LIVE}, which includes 82 live streams and nearly 250 hours of video. Based on the \textit{LIVE} dataset, to better evaluate the accuracy and efficiency of entity linking on video streams, we introduce a time-weighted decay metric named \textit{RoFA}, which comprehensively considers the accuracy and robustness of model predictions while also imposing requirements on online performance. Considering the \textit{OVEL} task and \textit{LIVE} dataset, as shown in figure \ref{FIntro} we analyze the \textit{OVEL} task, which poses several key challenges:

\textbf{Much Noise.\  }Real-time scenarios often exhibit a multitude of visual scenes and various sounds, which can introduce interference in entity recognition. For instance, in live-streaming e-commerce scenarios, hosts tend to use a significant number of interjections, engage in interactions with viewers, or interact with other hosts. These can cause substantial interference in the recognition of entities.

\textbf{Timeliness.\  }In online scenarios, which are characterized by strict time constraints, the prompt identification of salient entities and their timely recommendation to potential users often results in enhanced economic benefits. The expeditious identification of significant entities entails a challenging prerequisite for timeliness.

\textbf{Domain knowledge.\ }Recognizing certain products requires a certain level of domain knowledge, and individuals unfamiliar with the domain may struggle to make accurate identifications. For instance, it might be challenging for some people to distinguish the specific generation and specific superstar's basketball shoes.

Considering these challenges of \textit{OVEL} task, We propose several methodologies to address these challenges. Firstly, to address the issue of high noise levels in online scenarios, we propose adopting a LLM-based information extraction approach, aiming to extract information from videos that are more relevant to the entities.
Secondly, to address the issue of timeliness, we utilize a memory block to store information before the current inference moment. For the subsequent moment, only the information within the time interval and the memory block before this moment need to be inputted, ensuring real-time performance. And we delegate the management of the memory block to the LLM.
Furthermore, to tackle the domain-specific nature of live recognition, we propose utilizing a model retrieval to provide examples to LLM, enabling the LLM to possess a broader background knowledge. Lastly, when leveraging LLM for entity linking, a huge amount of entity candidates causes insufficient text length. We introduce a two-stage framework where MEL Methods act as candidate retrieval, and the LLM is used for entity disambiguation. This approach not only utilizes the capabilities of the large language model but also mitigates the issue of resource consumption.
In summary, the main contributions of this paper are as follows:

\begin{itemize}
\item {}To the best of our knowledge, we introduce the task of online entity linking(\textit{OVEL}) for the first time, focusing on improving the accuracy and efficiency of entity recognition in online videos.
\item {}Building upon live streaming scenarios, we have created a dataset for live stream product recognition, comprising 82 live stream videos, approximately 250 hours of video, and nearly 3,000 data instants. And a corresponding metric named \textit{RoFA}.
\item {}To better address the task of \textit{OVEL}, we propose a framework for the comprehensive management of video stream information based on LLM as a memory manager. Additionally, we leverage retrieval for LLM to manage memory better and employ a two-stage approach for entity linking. Subsequent experiments validate the effectiveness of our framework.
\end{itemize}

\section{Related Work}

\subsection{Multi-modal Entity Linking}
Multimodal Entity Linking (MEL) is an extension of entity linking that links mention in multi-modal information (e.g., images, audio, or videos) to a corresponding knowledge base. Existing research primarily focuses on static image-text pairs. Researchers\cite{adjali2020building,adjali2020multimodal,zhou2021weibo,wang2022wikidiverse,wang2022multimodal,gan2021multimodal,sun2022visual,chengmei2023mmel,xing2023drin,shi2023generative,yao2023ameli,zhang2021attention} constructed multiple datasets for different scenarios or proposed various multimodal representation methods, integrating features from different modalities to facilitate entity mention and entity matching. 

These studies primarily focus on static textual and graph data and have not been extended to the domain of videos. In the realm of entity linking in videos, \citet{li2015semantic} introduced a dataset for entity linking in videos and linked prominent entities from the videos to the knowledge base. For example, they linked highlights of Kobe Bryant's career to the entity ``Kobe Bryant''. \citet{venkitasubramanian2017entity} established a dataset for documentary video linking, utilizing video descriptions and content recognition to identify corresponding animals such as lions, birds, and others. These methods have two limitations. Firstly, the granularity of entities in videos is often too coarse, lacking fine-grained entity identification. Secondly, they primarily focus on pre-stored videos, linking them to the knowledge base with the whole video information, without considering real-time entity linking for online video streams.
\subsection{LLM as Memory Controller}
With the development of large language models (LLMs) \cite{devlin2019bert,radford2018improving,radford2019language,brown2020language}, LLMs that have been pre-trained on massive corpora have demonstrated remarkable capabilities\cite{ouyang2022training,wei2022emergent}. With the advent of powerful generative models such as GPT-4\cite{openai2023gpt4}, these models have demonstrated exceptional capabilities in generation, conversation, and the comprehension of human instructions, finding applications across a variety of downstream tasks. Recently, numerous researchers have integrated Memory with Large Language Models (LLMs), proposing frameworks to address resource constraints such as input length limitations inherent in LLMs. \citet{liang2023unleashing} proposed the utilization of memory to enhance the ability of LLMs to handle long texts, while \citet{zhong2023memorybank} introduced a customized memory mechanism specifically designed for LLMs. These frameworks offer potential for downstream applications of LLM-based agents. In this paper, we employ the LLM for memory block management and entity linking.
\subsection{Retrieval Augment Generation}
Despite the impressive capabilities demonstrated by models trained on large-scale corpora, they still suffer from phenomena such as hallucinations, long-tail problems, and knowledge decay. Retrieval augmentation, as a form of external corpora and knowledge enhancement, can alleviate these limitations of large models. In recent years, retrieval augmentation\cite{lewis2021retrievalaugmented,guu2020realm,lin2023radit} has been employed in various stages of model training, fine-tuning, and inference, leading to improved performance of models on downstream tasks. 

\cite{izacard2022atlas} utilized knowledge retrieval as a few-shot paradigm to enhance the performance of large language models in tasks such as knowledge question answering and fact-checking. \cite{vu2023freshllms} leveraged search engine retrieval to augment large language models, mitigating the issue of factual inaccuracies resulting from outdated knowledge. \cite{asai2023selfrag} improved the quality and factuality of model-generated outputs through retrieval and reflection. 
In this paper, we utilize retrieval augmentation to alleviate the issue of insufficient knowledge using LLM in domain-specific scenarios.



\section{Benchmark Construction and Evaluation}

\subsection{Problem Formulation}

Online Video Entity Linking (\textit{OVEL}) is a task designed for live video data streams. The goal of this task is to accurately identify salient entities in a live video stream, like the products highlighted by the anchor in the live broadcast scene. Given a 
live video, for instance, within the first 3 seconds, the host first mentions a specific pair of Nike shoes, followed by another 3 seconds of detailed introduction of it, then 3 seconds of answering questions from the live audience, and another 3 seconds of introduction to Nike shoes, and followed by a 3 seconds of Adidas’s competitive shoes. The prominent entities in these video streams should be Nike shoes, \textit{OVEL} should predict the Nike shoes for each 3 seconds input accuracy and robustness. This uneven distribution of information poses significant challenges to \textit{OVEL} task.


Considering mentioned above, The input of \textit{OVEL} should be a sequence of frames that accumulated with time. Given a live video $V_{m}$ consisted of a list of video clips ${V_{m}=\{v^1_{m},v^2_{m},...,v^t_{m},...,v^n_{m}\}}$, where ${v^t_{m}}$  represents the $t$-th clip of video ${V_{m}}$. And a predefined knowledge base ${KB=\{ e_1, e_2, ..., e_{j}\}}$, where each entity in the knowledge base has corresponding multimodal information. Below is the formal formulation of \textit{OVEL} at timestamp $t$:
\begin{equation}
  \mathop{\arg\max}\limits_{e_{p}^t}{P(e_p^t|[v^1_{m},v^2_{m},...,v^t_{m}],KB)}
\end{equation}


An entity $e_{p}^t$ should be predicted at each timestamp $t$ with the video information before timestamp $t$. Hence, a list of entities $\{e^1_{p},e^2_{p},...,e^t_{p},...,e^n_{p}\}$ will be predicted in the video $V_{m}$. Each entity in the prediction list should be as similar as ground truth $e_{m}$. This places lots of challenges on the robustness and accuracy of the algorithm.

\subsection{Dataset Construction and Analysis}
To advance the research on \textit{OVEL} task, we have built an e-commerce video stream entity linking dataset based on live streaming scenarios. The construction of the dataset consists of three main steps. Firstly, the initial raw videos and their corresponding multimodal knowledge base are obtained. The second step involves segmenting the corresponding live videos into data instances and manually annotating the entities in the knowledge base. The third step entails simulating online input by dividing each data instance into a list of video clips based on their playback time. The details of dataset construction and dataset Analysis can be found in Appendix~\ref{Adata}.

\subsection{Evaluation For OVEL} 
Evaluating the \textit{OVEL} task is not inherently straightforward and presents certain challenges. In the domain of live streaming, early identification of entities is increasingly effective for recommendation algorithms, potentially leading to greater economic benefits. The simplest approach involves assigning higher scores to instances where the correct location of real entities is identified earlier in the video. However, there is a possibility of correct recognition in the first minute but misidentification after one and a half minutes, which puts forward requirements for the robustness of the algorithm. Based on the aforementioned characteristics, we propose a comprehensive metric that considers accuracy, online performance, and robustness, we call it Robust online Fast Accuracy (RoFA). Below is the formulation of RoFA:

Given a list of prediction results in the temporal sequence $\{e^1_{p},e^2_{p},...,e^t_{p},...,e^n_{p}\}$, where the scores for predictions made later should be lower. Hence, we have devised a weighted decay mechanism that is proportional to the size of the prediction results. we initialize a linearly decreasing weight ${\{w_{0},w_{1},...,w_{t},...,w_{n}\}}$. For example, the weight of the first prediction is set to 1, and the weight of the last prediction is set to 0.2 (${w_{0}=1}$, ${w_{n}=0.2}$). The weights between these two windows decrease linearly, which aims to evaluate the fast and robust performance of algorithms.  As we only recommend the best matching product to users, so considering the prediction result for each video clip, if the prediction is correct, the score should be 1. Meanwhile, if the prediction is incorrect, the score is 0. The final metric is calculated as the sum of scores divided by the sum of weights, representing the average score. The calculation method of RoFA is as follows:
\begin{equation}
  \textit{RoFA}=\frac{\sum_{i=0}^{n} w_{i} \cdot score_{i}  }{\sum_{i=0}^{n} w_{i}}
\end{equation}
while ${score_{i}}$ is calculated as below, as ${e_{m}}$ denotes the ground truth of the video, and $e_{i}^p$ denotes the predicted entity.
\begin{equation}
  score_{i}=
\begin{cases}
    \text{1,} &  \text{if $(e_{i}^p=e_{m})$ } \\
    \text{0,} &  \text{if $(e_{i}^p \ne e_{m})$ } 
\end{cases}
\end{equation}

\section{Methods}
In this section, we will first present the overall framework of the methodology, followed by an introduction to the summary modules that constitute the methodology and an overview of the main components of the LLM as the memory controller. Finally, we will introduce the two-stage entity linking methods.
\subsection{Overview of the Framework}
\begin{figure*}
\centering
\includegraphics[width=\textwidth]{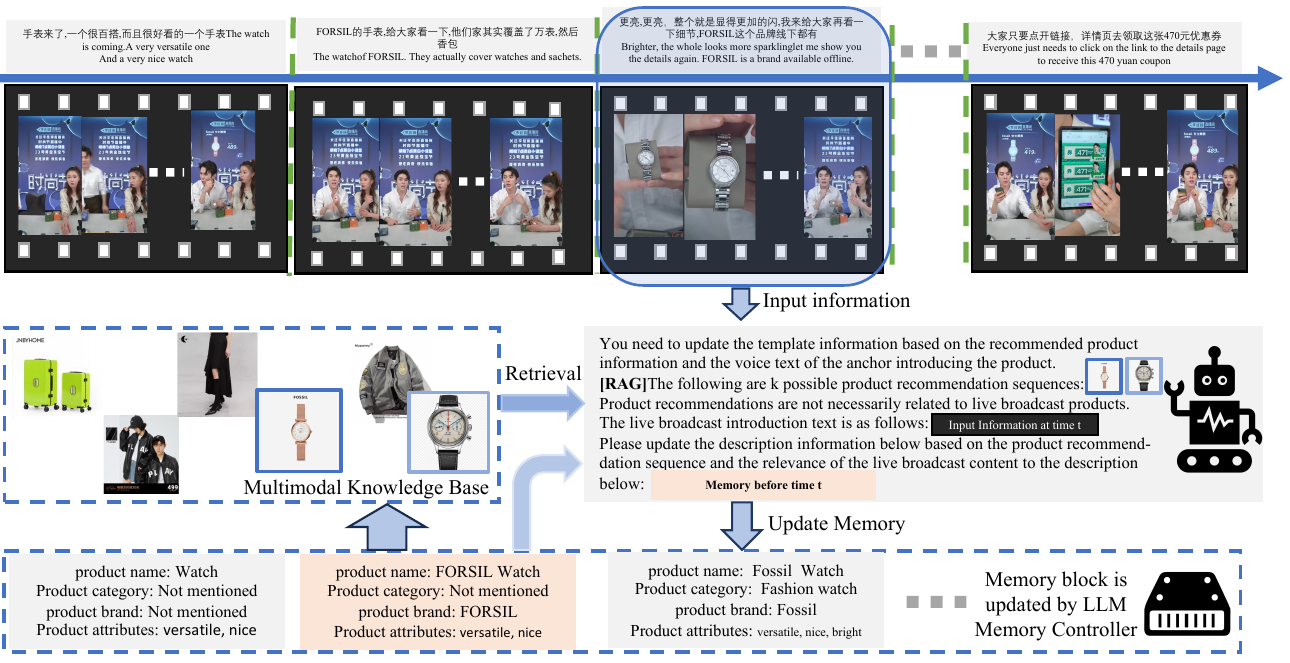}
\caption{Overview of framework structure. The initialized memory block is obtained through the summary module and used alongside keyframes extracted from the video by MEL to get initial retrieval candidates. At time t, the LLM memory controller acquires video information within the current input time interval, the memory block before time t, and incorporates retrieval results to update the content within the memory block.}
\label{Fmodel}
\end{figure*}

Figure \ref{Fmodel} illustrates the entire workflow of our Framework. When the input is an online video, we initialize the initial memory block using the summary module. 
Then we leverage the memory block and image information from video clips to perform the initial retrieval of candidate products. 
At each time t, the LLM manager gets the current video information, accesses the content within the memory, and refers to the results obtained from the retrieval model to make decisions and update the memory from the previous time step. To better use LLM's capacity, we also employed a two-stage entity linking method. First is the retrieval model to retrieve the candidate entities, and give candidates to LLM for fine-grained entity disambiguation. Below we will provide a detailed description of each module.

\subsection{Summary Module}
For a given input of video clips $\{v_{0},v_{1},...,v_{t},...\}$, while $v_{t}$ represents the video clip at time $t$, transcribed speech text $\{s_{0},s_{1},...,s_{t},...\}$, and keyframe sequences $\{i_{0},i_{1},...,i_{t},...\}$ over time. The task of OVEL is to predict ground truth entities at every moment as accurately as possible.
As a task of multimodal entity linking, the fundamental model should be a multimodal retrieval model. The multimodal retrieval model aims to maximize the similarity between real-time videos and their corresponding entities while minimizing the similarity between non-matching entities.  This can be represented by the following equation:
\begin{equation}
Embed_{v}=Encoder_{v}(V_{s}^{t},V_{t}^{i}) 
\end{equation}
\begin{equation}
    Embed_{e}=Encoder_{e}(e_{m}^t,e_{m}^i) 
\end{equation}
\begin{equation}
    e_{t}=\mathop{\arg\max}\limits_{e_{m}\in KB} Sim\left (Embed_{v},Embed_{e}\right )
\end{equation}
\begin{small}
\begin{equation}
\label{Ebase}
V_{s}^t=[s_{0}:s_{1}:...:s_{t}]\quad
V_{i}^t=[i_{0}:i_{1}:...:i_{t}]
\end{equation}
\end{small}

In the equation, $Sim$ denotes the similarity calculation, while $Encoder$ represents the encoder component for both the video and the entities in the knowledge base. The video contains two multi-modal information: speech text and images. $V_{t}$ contains all the information before time $t$.

However, in the context of live streaming, real-time videos present dynamic and evolving information, accompanied by a substantial amount of irrelevant noise, such as the host's habit of introducing ``all girls'' and engaging with the audience. To address this issue, we first propose an approach that leverages a LLM for extracting textual content from speech. Equation \ref{Ebase} is replaced with the following formulation:
\begin{equation}
V_{s}^t=LLM_{summary}([s_{0}:s_{1}:...:s_{t}])\quad
\label{Eextract}
\end{equation}
We utilize speech text following summaries for multimodal retrieval, which forms the summary module of our proposed method.


\subsection{Memory Controller Module}

However, online entity linking poses a challenge in terms of responsiveness. As the video progresses in time, we encounter a more important challenge. The length of the textual content extracted from speech increases over time, resulting in longer summaries time. At this point, using the summary module cannot meet the real-time requirements. To address this issue, we propose utilizing a memory block to store past extracted information.
As shown in the bottle of Figure \ref{Fmodel}, which aims to store past information with limited resources. The memory block module is designed to record entity-related attributes from previous video clips. When processing new video segments, only the current memory information needs to be updated, thereby avoiding linear growth in the number of tokens required for inference per clip. The input speech text in Equation \ref{Ebase} is replaced by the equation listed below:
\begin{equation} 
V_{s}^t=Mem_{t}=LLM(s_{t},Mem_{t-1})\quad
\label{Ememory}
\end{equation}
From the equation, it can be observed that at each time step, only the memory from the previous time step and the textual information of the current clip are required as inputs. 

However, in the live-streaming scenario, there are limitations. The granularity of products in live streaming is relatively fine, requiring domain-specific knowledge. Additionally, there is a significant amount of irrelevant information present in the videos. If we solely rely on an LLM trained in a general domain to manage the memory block, there is a risk of extracting a large amount of irrelevant information. To ensure that the memory block is primarily filled with information related to the products, we combine it with the retrieval model. The products obtained through multimodal retrieval are simultaneously considered by the LLM, which acts as guidance for better memory block management. 
Equation \ref{Ememory} is replaced by the following formula:
\begin{equation} 
V_{s}^t=Mem_{t}=LLM(s_{t},Mem_{t-1},[E_{k}])
\end{equation}
\begin{equation} 
[E_{k}]=Top_{k}(\mathop{\arg\max}\limits_{e_{m}\in KB} Sim\left (Embed_{v},Embed_{e}\right ))
\label{EtopK}
\end{equation}
In equation \ref{EtopK}, $Embed_{x}$ denotes the embedding encoded by corresponding encoders. From the equation, it can be observed that at each time step, the inputs consist of the memory from the previous time step, the textual information of the current slice, and the retrieval results from the retrieval model. This not only fulfills the requirements of real-time inference but also alleviates the issue of insufficient domain-specific knowledge in LLM.
\subsection{Two-stage Entity Linking}
The LLM demonstrates remarkable capability, which we desire to use for entity linking. However, in real-time scenarios, it is challenging to provide all the candidate entities to the LLM due to its limited context length, and fine-grained non-deterministic generation is also difficult. Drawing from previous approaches\cite{wang2022wikidiverse}, we divide the linking process into two steps: the first step involves the retrieval model to get entity candidates, and the second step involves the entity disambiguation made by the powerful LLM. 
The formula for this progress is illustrated in the following: 
\begin{equation} 
e_{k}=\mathop{\arg\max}\limits_{e_{m}\in KB} Sim\left (Embed_{v},Embed_{e}\right )
\label{EentK}
\end{equation}
\begin{equation}
e_{t}^p=LLM_{choice}([e_{k}^1,e_{k}^2,...,e_{k}^n]) 
\end{equation}
From the formula, it can be observed that initially, a reduced set of entity candidates is retrieved using MEL. Then, LLM is employed to select the optimal candidate entity from this set. This approach not only leverages the powerful background knowledge of LLM but also reduces the time-consuming inference capacity. 

Above is the comprehensive presentation of our proposed framework. The following experiments show that our method ensures real-time performance while effectively enhancing overall performance. 
\section{Experiments}
In this section, we primarily focus on analyzing the experimental results conducted on the constructed dataset. Firstly, we present the main experimental results of the overall framework. Secondly, we examine the comparative analysis of different approaches in terms of temporal performance. Lastly, we compare the performance of traditional metrics on static videos. Furthermore, several intriguing phenomena emerged during our experimental process, which can be observed in Appendix \ref{Aexp}.
\subsection{Experiments setting}
\textbf{Model selection.} We employed Qwen-14B-Chat\cite{bai2023qwen} as the LLM. We also utilized powerful models such as ChatGPT, However, due to a lack of training data in the e-commerce domain, these models performed poorly on the task. For the retrieval model, we selected Chinese-CLIP\cite{yang2023chinese} and AltCLIP\cite{chen2022altclip} for multimodal retrieval. We also attempted to translate the Chinese memory block and product database into English and used English text-image retrieval, however, due to the transcription errors in the text converted from voice in the live streaming scenario, including brand names, it is challenging to achieve fine-grained product recognition. 
Our static experimental results also demonstrate the underperformance of other models on the dataset.
Therefore, we only present the results obtained with Chinese-CLIP and AltCLIP.  Chinese-CLIP involves fine-tuning a well-trained Chinese text encoder and image encoder for high-quality text-image retrieval through contrastive learning. AltCLIP, on the other hand, incorporates Chinese training data into CLIP, making it a multilingual text encoder model. We employed Chinese-CLIP architectures based on ResNet\cite{he2015deep} and RBT3, as well as ViT-H/14\cite{dosovitskiy2021image} and RoBerta\cite{liu2019roberta}. These architectures are denoted as CN-CLIP\_B and CN-CLIP\_L, respectively. AltCLIP utilized official weight initialization.

\textbf{Implement Details.} For the method proposed in this article is designed for online performance analysis, all experiments are performed on the same machine. Our local machine has four 3090 GPUs. To facilitate better inference, we deployed Qwen-14B-Chat on an A100 80G machine and used API calls to manage memory blocks through the LLM controller. Due to limitations in local inference memory, we randomly sampled a product database approximately 10 times larger than the test set from the knowledge base. We fixed this subset of 3,000 products as the candidate pool, and the test set consisted of 275 video samples. We assume that the model begins generating outputs after processing 10 video clips, indicating that the model starts linking from the 10th video clip. To better utilize the sequential information in memory, except for the Base method, all other approaches perform inference once every 5 video clip sizes. The inference results are then replicated for all five video clips. All methods are finetuned on the training set.
\subsection{Main result}
In this section, we added our framework to two multi-modal retrieval models.  It is worth mentioning that on the test set, AltCLIP performed poorly in retrieval due to the main language of pre-trained data being English and a lack of domain knowledge. On the other hand, CN-CLIP\_B had a smaller parameter size and its performance was not as good as CN-CLIP\_L. To compare the effectiveness of different modules, we denote the model that directly employs multimodal retrieval as ``Base'', and our proposed LLM as memory controller as ``Ours''. The RoFA results are presented in the Table \ref{Tmain}.
\begin{table}[h]
\centering
\scalebox{0.9}{
\begin{tabular}{c|ccc}
    \hline
    method & AltCLIP  & CN-CLIP\_B & CN-CLIP\_L \\
    \hline
    $Base$ & 2.32 & 23.16 &  36.68\\
    \hline
    $Ours_{-M}$ & 4.80 & 42.17 &  56.60  \\
    $Ours_{-R}$ & 4.85 & 35.30 &  47.02 \\
    $Ours$ & \textbf{13.20} & \textbf{48.16} & \textbf{60.20}  \\
    \hline
\end{tabular}}
\caption{RoFA results of proposed methods.While $Ours_{-R}$ represents the removal of the retrieval module, while $Ours_{-M}$ represents the removal of the memory block module.}
\label{Tmain}
\end{table}
From the table, it can be observed that the approach combining retrieval model retrieval with LLM achieved the highest performance. Particularly, our method combining the CN-CLIP\_L model achieved the best results, likely due to CN-CLIP\_L's superior performance on the test sets compared to the other two retrieval models. Additionally, our method exhibited the most significant improvement on AltCLIP, reaching nearly 300\%. We speculate that this is because while AltCLIP performs poorly in retrieval alone when we divide the task into two steps and provide sufficient candidate options to the LLM, the LLM can often select the best candidates. In most cases, using a single memory management approach yields slightly inferior results compared to using full summaries, as the structure of memory management lacks complete information, leading to some information loss. The results also demonstrate that our method provides substantial improvements when the retrieval model performs poorly. This suggests that in low-resource scenarios where the retrieval model lacks training data, leveraging the combination of the LLM can serve as a viable solution.
\subsection{Online performance analysis}
\begin{figure}[h]
\centering
\includegraphics[width=0.5\textwidth]{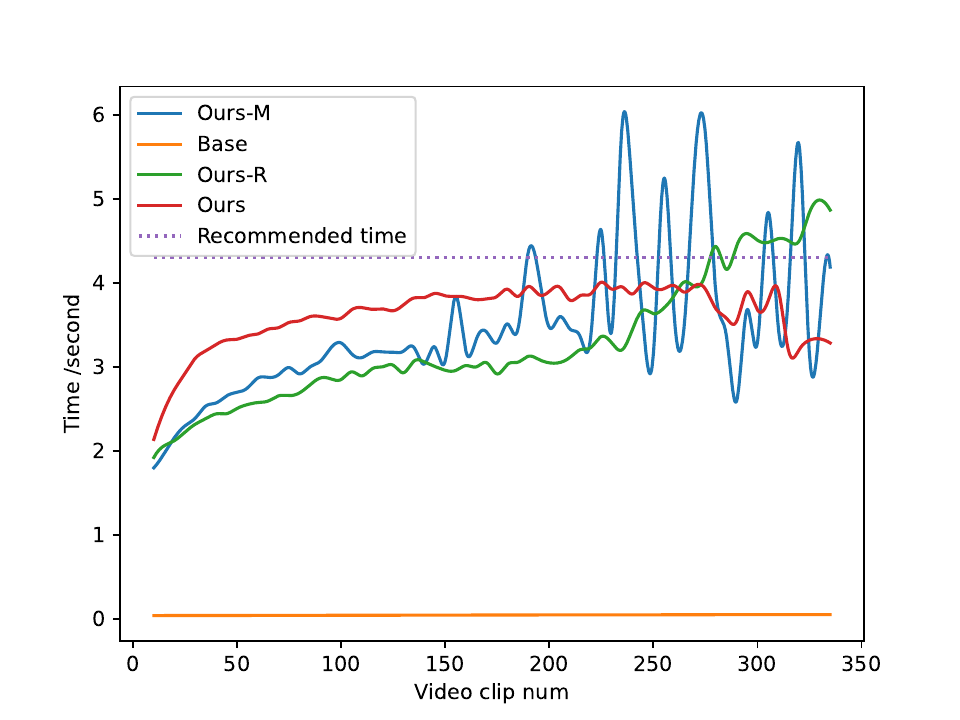}
\caption{Inference time of different method. The Recommended time is determined based on the optimal inference time consumption provided by the actual application scenario.}
\label{FReal}
\end{figure}

In this section, we analyze the online performance of different methods. We evaluated the time performance of various methods on the test set. We assessed the time taken to give a predicted entity and recorded the inference time on the test set at intervals of every five video clips. After calculating the average time for each method, the smoothed results are presented in Figure \ref{FReal}.
From Figure \ref{FReal}, it can be observed that ``Base'' utilizes the retrieval model and yields the best time performance. The time cost of the Ours-M and Our Ours-R significantly increases as the number of video clips grows. It needs to be mentioned that when the window size exceeds 200, these methods surpass the recommended inference time, thereby potentially failing to provide meaningful linked entities within the given time interval. On the opposite, Our method initially exhibits a rapid increase in time cost, followed by a tendency toward stability.

Analyzing the reasons behind this situation: as the number of video clips increases, the length of the memory block also increases, resulting in an information increase in all methods. In the later stages of inference, due to its domain knowledge from the retrieval model, Ours method tends to have content that is more related to specific products and remains fixed. On the other hand, the Ours-R may continue to accumulate irrelevant information as it lacks related knowledge. And Ours-M method exhibits some instability due to variations in the length of text in different video clips, the reason may be the lack of a complete memory, the extracted information may be inconsistent in format, and there may be insufficient or redundant.

\subsection{Static experiment results}
To compare with some of the existing methods for Multimodel Entity Linking (MEL) and Video Entity Linking (VEL), we treat each video as an individual data instant and perform video entity linking across the entirety of the video's content to ascertain the applicability of our method to static data as well. In this chapter, we have selected a variety of representative approaches for evaluation. These include CLIP4Clip\cite{luo2022clip4clip} in the domain of video retrieval, a purely textual entity linking approach BLINK\cite{logeswaran2019zero}, the multimodal entity linking method V2VTEL\cite{sun2022visual}, and other multimodal retrieval methods such as AltCLIP and Chinese-CLIP. The experimental metrics primarily utilized are Recall and Mean Reciprocal Rank (MRR) at K. The experimental outcomes are as exhibited in the Table \ref{Tstatic}.
\begin{table}[h]
    \centering
    \scalebox{0.9}{
    \begin{tabular}{c|cccc}
        \hline
        \textbf{Method} & R@1  & R@5 & MRR@3 & MRR@5\\
        \hline
        CLIP4clip & 1.06 & 8.05 &  2.14 & 3.10\\
        AltCLIP & 8.95 & 20.62 &  12.4 & 13.3 \\
        V2VTEL & 9.09 & 24.1 &  13.0 & 14.2 \\
        BLINK & 42.2 & 72.7 &  53.7 &  54.8\\
        CN-CLIP & 55.1 & 75.3 & 62.2 & 63.2 \\
        \hline
        Ours & \textbf{57.7} & \textbf{82.3} & \textbf{66.0} & \textbf{66.8} \\
        \hline
    \end{tabular}}
    \caption{Static results of proposed methods.}
    \label{Tstatic}
\end{table}

From the table \ref{Tstatic}, it can be observed that our method achieves the best performance, once again demonstrating the effectiveness of our approach. Furthermore, the performance of the CLIP4clip and V2TVEL approaches compared to pure text-based BLINK is poor, indicating that text plays a more significant role in our scenario. Among the proposed methods, only CN-CLIP and AltCLIP incorporate multimodal inputs, and they exhibit favorable results, which is why we have chosen them as our multimodal retrieval models.
 

\section{Conclusion}
In this paper, we propose an Online Video Entity Linking (\textit{OVEL}) task for online videos, construct the \textit{LIVE} dataset based on live streaming scenarios, and introduce the RoFA metric, which considers robustness, timeliness, and accuracy. Based on the dataset, we present a method that combines LLM with a retrieval model for memory management, which handles the \textit{OVEL} task efficiently. Experimental results demonstrate the effectiveness of our approach. However, \textit{OVEL} is a highly challenging task, and we earnestly invite researchers to join us in the field of online video entity linking. 


\bibliography{anthology,custom}
\bibliographystyle{acl_natbib}

\appendix

\section{Dataset Construction and Analysis}
\label{Adata}
\begin{figure*}[ht]
\centering
\includegraphics[width=\textwidth]{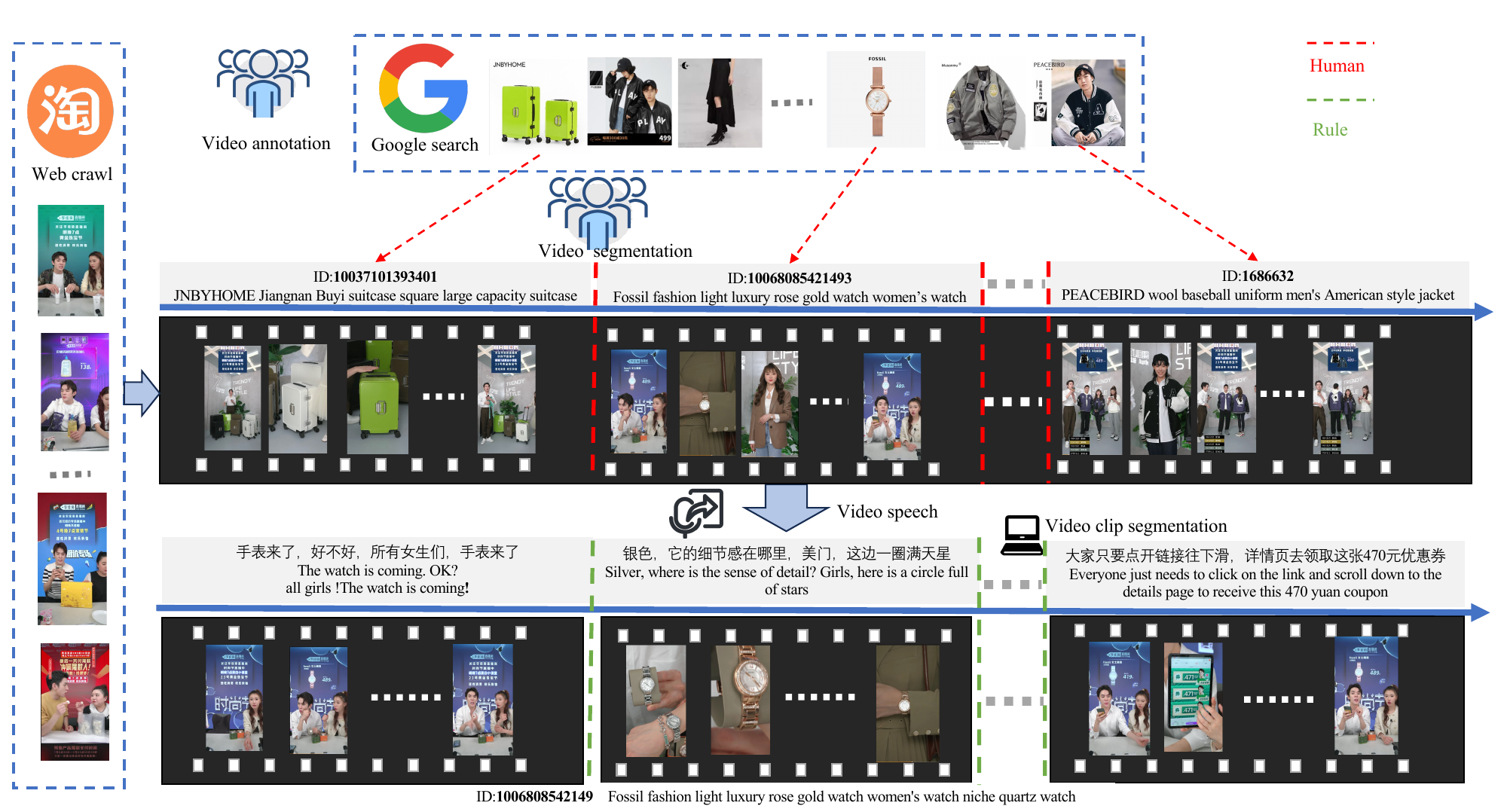}
  \caption{The procedure of LIVE dataset construction.}
  \label{FData}
\end{figure*}
\subsection{Dataset construction}
As shown on the left side of Figure~\ref{FData}, we first obtained the original files of 82 live stream videos from Taobao live\footnote{https://taolive.taobao.com/}. On average, each livestream video had a duration of 5.6 hours and included an average of 51.3 live product items. While crawling the videos, we also obtained a list of the names of the products featured in each live stream, as shown lower left of Figure~\ref{FData}. 

However, the order of the products may not correspond directly. We need to complete video segmentation and annotate the 
corresponding products in those segments. We hired five data annotators who followed a unified standard for annotation. On average, each annotator spent two days on the task. Additionally, two skillful individuals involved in the project reviewed and corrected the annotations for quality assurance. After completing the video segmentation and product annotation, we needed to retrieve corresponding images for the products using the product names above. We employed a combination of rule-based retrieval and manual inspection to gather product images. Initially, we conducted a Google Image search\footnote{https://www.google.com/search} using the names of the products. Firstly, we filtered the search results based on prominent Chinese e-commerce domain names (such as www.taobao.com, www.jd.com, and so on). Besides, we prioritized the results based on the semantic similarity between the search results and product names. We intercepted the top ten results after sorting. Finally, we manually selected the most suitable product image from the top 10 candidate products as the completion of the image information for the knowledge base. The procedure of this step is shown in the middle of Figure~\ref{FData}.

And finally, to facilitate real-time input, we divided the video into video clips. Previous research has shown that in fine-grained entity linking, such as ``Nike Jordan 36th Generation High-Top Basketball Shoes'', textual information plays a more significant role in identification. Therefore,  to better process the text from the video speech, we utilized OpenAI's Whisper \cite{radford2022robust} model to transcribe the speech in the video. The video is sliced according to the sentence segmentation results. That is, each sentence corresponds to the smallest video slice, ultimately creating a simulated real-time video input. The procedure of this step is shown in the right of Figure~\ref{FData}.
\subsection{Dataset analysis}
\begin{table}[h]
\centering
\scalebox{0.9}{
  \begin{tabular}{c|ccccc}
  \hline
    Dataset& $V_{num}$&$M_{Ave}$ &$C_{Ave}$&$S_{Ave}$&$E_{num}$\\
    \hline
    \textit{LIVE} & 2870 & 4.66 & 155.33 & 3.78 & 23519 \\
    \hline
\end{tabular}
}
\caption{\label{TData}Overview of LIVE dataset. While $V_{num}$ represents the number of video instances, $M_{Ave}$ represents the average minutes per instance, $C_{Ave}$ represents the average number of video clips per instance, $S_{Ave}$ represents the average seconds of a single clip data, and $E_{num}$ represents the number of entities included in the knowledge base.} 
\end{table}
\textbf{Basic analysis}
After dataset construction, we obtained 2780 data instances, each corresponding to an entity. The average duration of each data instance is 279.58 seconds, and on average, there are 155.33 video clips per data instance. 
In addition, for the auxiliary samples, we selected different products of various brands from the Google search results to create hard negative samples for distinguishing between products. This resulted in a total of 19653 auxiliary samples.
Table\ref{TData} presents the basic information of the dataset.

\begin{figure}[h]
\centering
{\includegraphics[scale=0.45,width=0.5\textwidth]{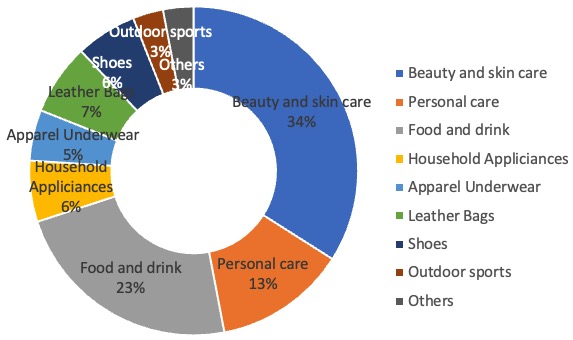}}
\caption{Overview of entity distribution.}
\label{FED}
\end{figure}

\textbf{Entity distribution}
The product catalog contains numerous categories, such as beauty and personal care, outdoor sports, food, and beverages, as shown in Figure~\ref{FED}, which displays the distribution of entities in the product catalog, with the beauty and personal care category being the most prevalent. Fine-grained analysis of entities in the product catalog involves distinguishing between similar products. For instance, the Nike Jordan 37th generation high-top basketball shoes, the Nike Air Force 1, and the Under Armour Curry 11 basketball shoes involve the design of the same product across different brands, as well as different products within the same brand. The \textit{OVEL} task aims to identify specific brands and models of the entities, which further emphasizes the difficulty of this task.
\section{Memory Block Analysis}
\label{Aexp}
Memory is a very important module proposed in this paper, and its format and management form are also particularly important. In the next two sections, we will discuss the impact of memory format and memory management on experimental results.
\subsection{Memory format analysis}
The memory block primarily stores attributes related to commodities, such as brand, category, etc. However, determining how to store these attributes is a crucial issue. We have opted for structured, semi-structured, human, and model-generated summaries as the forms of storage. Apart from using different prompts during initialization, the same prompts are used for the process of memory updating. The experimental results are presented in Table \ref{Tform}.
\begin{table}[h]
\centering
\scalebox{0.9}{
  \begin{tabular}{c|cccc}
  \hline
      Format & Struct & Semi-struct & Human & LLM \\
    \hline
    RoFA & \textbf{60.20} & 50.94 & 57.72 & 59.36\\
    \hline
    \end{tabular}}
    \caption{RoFA results of different memory formats.} 
    \label{Tform}
\end{table}\\
From the table, it can be observed that the prompt in the form of Struct yields the best performance, followed by the model-generated results. The Human storage method, which bears similarity to the self-generated structure by LLM (Large Language Model), exhibits inferior performance. The least effective approach is the Semi-struct method. This is because we treat the OVEL task as an extraction task, and the struct data represented in tuple form may be more suitable for such tasks. LLM demonstrates a good understanding of the data it generates, and the human storage method, similar to LLM, also exhibits decent performance.

Upon analyzing the Semi-struct approach, we found that it only contains ``commodity name: commodity attribute:'' forms. This has a higher probability of being influenced by the structure of the recommended reference name by the small model. This issue can be addressed by using better prompts. Additionally, some crucial attributes such as brand and category are placed within the attributes, making them less prominent and resulting in suboptimal performance.
\subsection{Memory management}
The management of memory blocks is a crucial aspect discussed in this paper, which can be observed in several key aspects. Firstly, in the context of live streaming, simply relying on agent management of memory over long time windows may result in drift and a gradual deviation over time. The small model provides references for the large model, and the large model tends to excessively rely on information from the small model, causing the memory to be associated with negative samples. Consequently, the retrieval results of the small model in the next iteration deviate, leading to an increasing deviation over time. As shown in the table below, the correct sample is ``Wufangzhai egg yolk'', but the first instance excessively relies on the small model, resulting in a biased outcome. In contrast, the second instance avoids such errors by the well-designed prompt. Therefore, it is advised to add error samples to demonstrations when using small models to help LLMs.
\begin{figure}[h]
\centering
\includegraphics[width=0.5\textwidth]{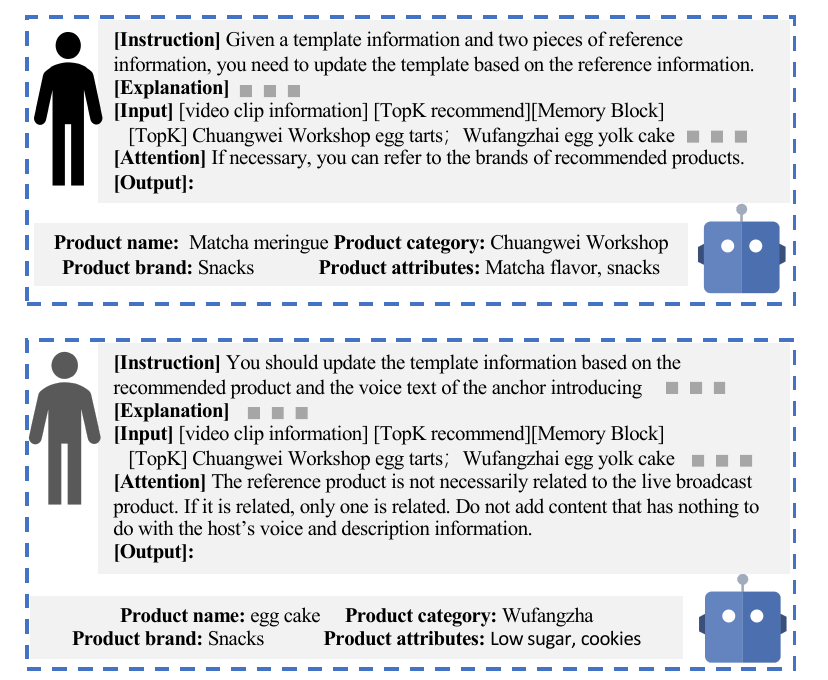}
\caption{Different prompt for Memory Management.}
\label{Fmem}
\end{figure}
\end{document}